\documentclass[10pt,twocolumn,letterpaper]{article}

\usepackage{cvpr}
\usepackage{times}
\usepackage{epsfig}
\usepackage{graphicx}
\usepackage{amsmath}
\usepackage{amssymb}

\usepackage[breaklinks=true,bookmarks=false]{hyperref}

\cvprfinalcopy 

\def\cvprPaperID{****} 

\begin{document}

\title{Surveillance Video Parsing with Single Frame Supervision}
	
	\author{Si Liu$^1$, Changhu Wang$^{2}$, Ruihe Qian$^{1}$, Han Yu$^1$,  Renda Bao$^1$ \\
		$^1$ Institute of Information Engineering, Chinese Academy of Sciences\\
		$^2$Microsoft Research 		Beijing, China, 100080\\
		\{liusi, qianruihe, yuhan\}@iie.ac.cn, chw@microsoft.com,roger\_bao@163.com}


\maketitle

	\begin{abstract}
		Surveillance video parsing, which segments the video frames into several labels, e.g.,   face, pants, left-leg, has wide applications \cite{zhao2014learning,deng2014pedestrian}. However,   pixel-wisely annotating all frames is tedious and inefficient.  In this paper, we develop a Single frame Video Parsing (SVP) method which  requires only one labeled frame per video in training stage.
		To parse one particular frame, the video segment preceding the frame is jointly considered. SVP \textbf{(\romannumeral1)} roughly parses the frames within the  video segment,  \textbf{(\romannumeral2)} estimates the optical flow between frames and \textbf{(\romannumeral3)} fuses the rough parsing results  warped by optical flow to produce the refined parsing result.  The three components of SVP, namely frame parsing, optical flow estimation and temporal fusion are integrated  in an end-to-end manner. Experimental results on two surveillance video datasets	show the superiority of SVP over state-of-the-arts.

	\end{abstract}
	
	\section{Introduction}

	In recent years, human parsing \cite{xiaodaniccv}  is receiving increasing  owning to its wide applications,  such as person re-identification  \cite{li2014deepreid,zhao2014learning} and person attribute prediction \cite{deng2014pedestrian}. Most existing human parsing methods \cite{liang2015deep,xiaodaniccv,yamaguchi2013paper}  target at segmenting the human-centric images  in  the fashion blogs.
	Different from  fashion images, parsing surveillance videos is much more challenging    due to the lack of labeled data.  It is very   tedious and time-consuming to annotate all the frames of a video, for a  surveillance video usually contains tens of thousands of  frames per second.

\begin{figure}[t]
		\begin{center}
			\includegraphics[width=1\linewidth]{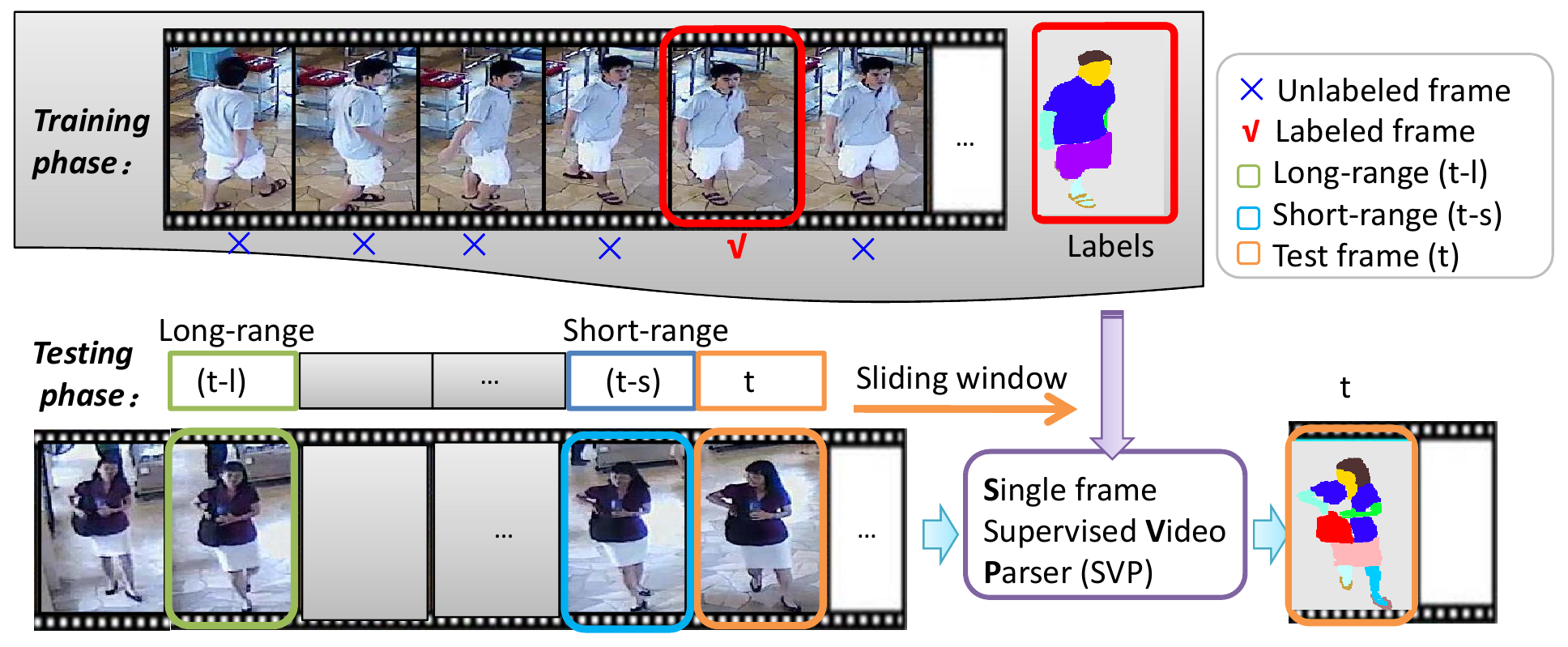}
		\end{center}
		\caption{During training, only a single frame per video (red check mark) is labeled, while others (blue x mark) are unlabeled.
			A  Single frame supervised video Parsing (SVP) network is learned from the extremely sparsely labeled videos. During  testing, a parsing window is slided along the video.
			The parsing result of testing frame $I_t$ (orange  box) is  determined by itself,  the long-range frame $I_{t-l}$ (green box) and the  short-range frame  frame $I_{t-s}$  (blue  box).  For better viewing of all
			figures in this paper, please see original zoomed-in color pdf file.}
		\label{fig:first_fig}
		\vspace{-5mm}
	\end{figure}

	In this paper, we target at an important, practically applicable yet  rarely studied  problem: \emph{how to leverage  the very limited labels to obtain a robust surveillance video parsor}?  More specifically, we mainly consider 	 an extreme situation, i.e., only one frame in each training video is annotated.
	Note that labeling is unnecessary in testing phase.  As shown in Figure \ref{fig:first_fig}, the  labeled frame per training video (red bounding box) is fed into the proposed Single frame supervised Video Parsing  (\textbf{SVP}) network. Insufficient labeled data always lead to  over-fitting, especially in the deep learning based method. The rich temporal context among video frames  can partially solve this problem.  By building the \emph{dense correspondences}, i.e., optical flow, among video frames, the single labeled frame can be viewed  as the {seed} to indirectly expand (propagate) to the whole video.  Most state-of-the-art optical flow estimation methods, such as EpicFlow \cite{revaud2015epicflow}, DeepFlow \cite{weinzaepfel2013deepflow}, LDOF \cite{weinzaepfel2013deepflow} etc,  suffer from relatively slow speed. Because the video parsing task requires extensive online  optical flow computation, a  real-time, accurate optical flow estimation is essential. Thus, it is a challenging but essential problem to build an end-to-end, efficient  video parsing framework by only  utilizing the  limited (e.g, only one) labeled images and  large amount of unlabeled images with online estimated dense correspondences among them.

	To tackle these challenges,  we propose the SVP network. 	As shown in Figure \ref{fig:first_fig},	to parse a test frame $I_t$, a parsing window
	which contains  $I_t$ and  several frames preceding it $\{I_{t-k}, k=0,...,l\}$,  is slided along the video.  Considering the computation burden and cross-frame redundancies, 	a triplet $\left\{ {{I_{t - l}},{I_{t - s}},{I_t}} \right\}$ is selected to represent the sliding window.  The \emph{long-range frame}   $I_{t-l}$ lies $l$ frames ahead of $I_t$ while
	\emph{short-range frame}  $I_{t-s}$ lies $s$ frames ahead of $I_t$.  Usually, $l>s$. 	
	They complement each other in that the short-range optical flows are more accurate, while the long-range  frames bring more  diversities.  The triplet is fed into SVP to collaboratively produce the parsing result.

	SVP contains three sub-networks. The image parsing sub-network parses the three frames respectively, while the  optical flow estimation sub-network builds the cross-frame pixel-wise correspondences.  In order to 	decrease  the interference of imperfect optical flow, a pixel-wise confidence map is calculated based on the appearance differences between  one image and its counterpart wrapped from the other image. 	Based on  the mined correspondences and their confidences, the temporal fusion sub-network fuses the parsing results of the each frame, and then outputs the final parsing result. Extensive experiments in the newly collected indoor and outdoor datasets show the superior performance of SVP than the state-of-the-arts.

	The contributions of this work are summarized as follows.
	\textbf{(\romannumeral1)} To the best of our knowledge, it is the first attempt  to  segment the human parts in the surveillance video by labeling  single frame  per training video. It has extensive application prospect.
	\textbf{(\romannumeral2)} The proposed  SVP framework is end-to-end  and thus very applicable for real usage. Moreover, the feature learning,   pixelwise classification, correspondence mining and the temporal  fusion are updated in a unified optimization process and collaboratively contribute to the parsing results.
	\textbf{(\romannumeral3)}  We will release  the collected indoor and outdoor video parsing dataset, which is expected to server as a benchmark for the further studies.

	\section{Related Work}
	
	\textbf{Image,  video and part semantic segmentation:} Long \emph{et al.} \cite{long2014fully}   build a FCN that take input of arbitrary size and produce correspondingly-sized output.
	Chen \emph{et al.} \cite{chen2014semantic} introduce
	‘atrous convolution’ in dense prediction tasks   to effectively enlarge the field of
	view of filters to incorporate larger context without increasing the number of parameters.
	Dai \emph{et al.} \cite{dai2015convolutional}  exploit shape information  via masking convolutional features.
	Hyeonwoo \emph{et al.} \cite{noh2015learning} propose Deconvolution Network for Semantic Segmentation   to identify detailed structures and handles objects in multiple scales naturally.
	Shelhamer \emph{et al.} \cite{shelhamer2016clockwork}  define a clockwork fully convolutional network for video semantic segmentation.   Fragkiadaki   \emph{et al.} \cite{fragkiadaki2015learning}  segment moving objects in videos by multiple segment proposal generation and ranking.
	
	{Part segmentation} is more challenging since  the object regions are smaller. For human parsing, Yamaguchi \emph{et al.}     \cite{yamaguchi2013paper}  tackle the clothing parsing  problem using a retrieval based approach. Luo \emph{et al.} \cite{luo2013pedestrian}  propose a  Deep Decompositional Network  for parsing pedestrian images into semantic regions.  Liang \emph{et al.} \cite{liang2015deep} formulate the human parsing task as an  Active Template Regression  problem.  Liang \emph{et al.} \cite{xiaodaniccv} propose a Contextualized Convolutional Neural Network to tackle the problem and achieve very impressing results.  Xia \emph{et al.} \cite{xia2015zoom}  propose the ``Auto-Zoom Net'' for human paring. Some other works explore how to jointly object and part segmentation using deep learned potentials \cite{wang2015joint}.  Although great success achieved, these methods can not be directly applied in our  setting where only one labeled frame per training video is available. In contract, SVP can effectively explore the extremely few labeled data and a large amount of unlabeled data.

	
	
	%
	%
	

	%
	%


	\textbf{Weakly/semi-supervised semantic segmentation:} Chen \emph{et al.} \cite{papandreou2015weakly} develop Expectation-Maximization (EM) methods to solve the  semantic image segmentation from either weakly annotated training data or  a combination of few strongly labeled and many weakly labeled images.
	Dai \emph{et al.} \cite{dai2015boxsup}  propose a method called ``Boxsup'' which  only requires easily
	obtained bounding box annotations. Dai \emph{et al.} \cite{lin2016scribblesup}   develop an algorithm to train convolutional networks for semantic segmentation supervised by scribbles.
	Bearman \emph{et al.} \cite{bearman2015s}   incorporate the point supervision
	along with a novel objectness potential in the training loss function of a
	CNN model. Xu \emph{et al.} \cite{xu2015learning} propose a unified approach that incorporates
	image level tags, bounding boxes, and partial labels to produce a pixel-wise labeling.
	Hong \emph{et al.} \cite{hong2015decoupled} decouple classification and segmentation, and learn a separate network for each task  based on the training data with image-level and pixel-wise class labels, respectively. These methods have achieved competitive accuracy in weakly/semi supervised semantic segmentation but are not designed for video parsing task.
	
	\begin{figure*}[t]
		\begin{center}
			\includegraphics[width=0.9\linewidth]{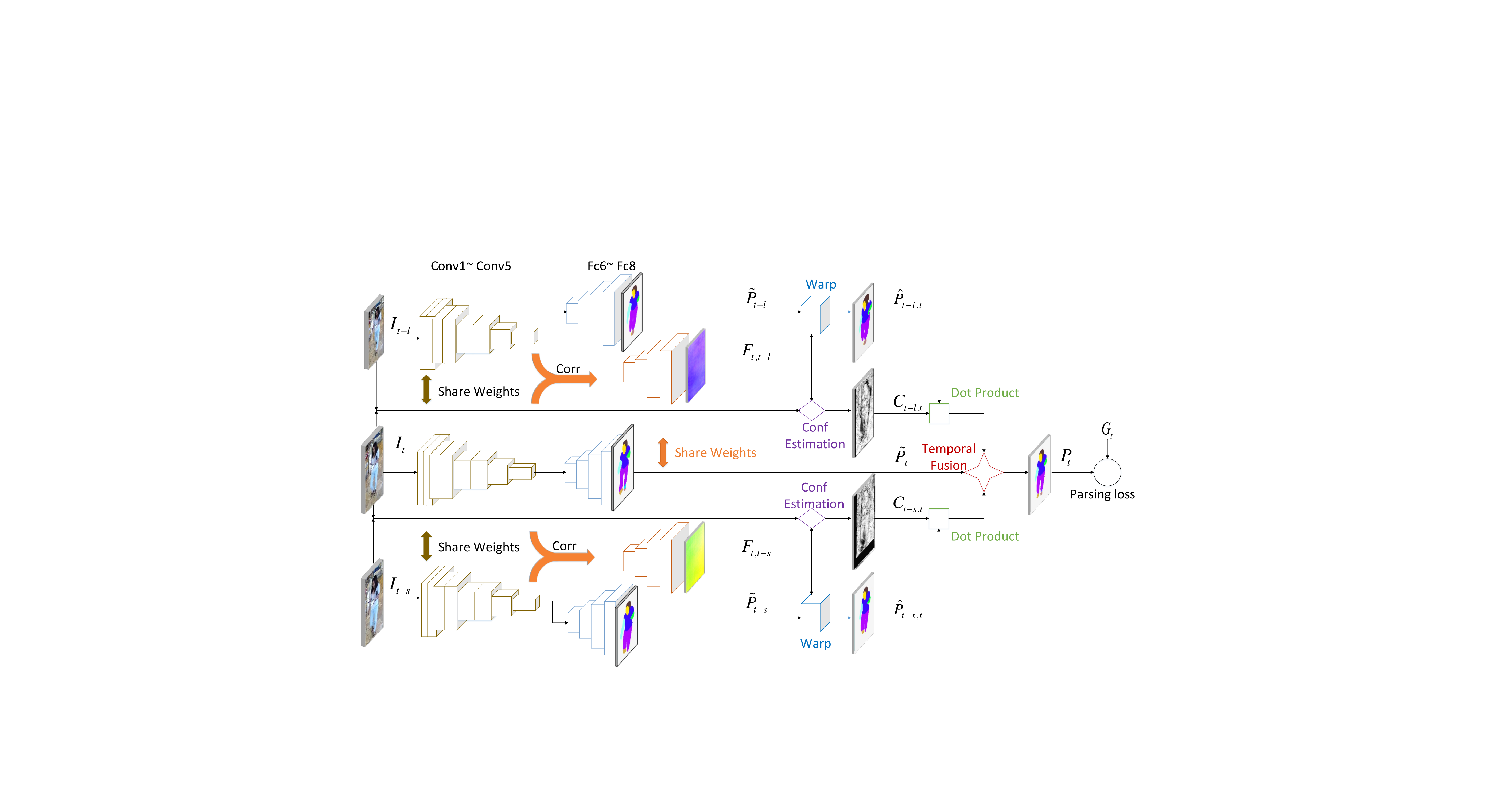}
		\end{center}
			\vspace{-4mm}
		\caption{The proposed  single frame supervised video paring (SVP) network. The network is  trained end-to-end.	}
		\label{fig:framework}
				\vspace{-4mm}
	\end{figure*}

	\textbf{Optical flow v.s.   semantic segmentation}:
	Sevilla-Lara \emph{et al.} \cite{sevilla2016optical}    segment a scene into things, planes,   and stuff and then pose the flow estimation problem using localized layers.  Bai \emph{et al.} \cite{bai2016exploiting}   estimate the traffic participants using instance-level segmentation. The epipolar constraints  is then used on each  participant to  govern each independent motion.  In these methods,  optical flow estimation benefits from semantic segmentation. However, SVP utilizes optical flow for better video parsing.

	Pfister  \emph{et al.} \cite{pfister2015flowing}  investigate a video pose estimation architecture that is able to benefit from temporal context by combining information across the multiple frames using optical flow.
	The key differences  is that the optical flow is estimated offlined using dense optical flow \cite{weinzaepfel2013deepflow} while DVP is an end-to-end framework.

	\section{Approach}
	
	In this section, we  first introduce the  SVP network. Then we will elaborate three sub-networks sequentially.

	\subsection{Framework}

	Suppose that we have a  video $\mathcal{V} = \{ {I_1}, \cdots ,{I_N}\}$, where $N$ is the number of frames.  The single labeled frame is  ${I_t}$, and  its corresponding groundtruth  is ${G_t}$.
	The pixel $j$ of the labelmap ${P_t}$ is  denoted as $P_t^j$  and  takes the value within the  range  $[1,K]$, where $K$ is the number of labels, such as ``face'', ``bag'' and ``background''.

	The SVP network is shown in Figure \ref{fig:framework}.
	The input is a triplet $\{ {I_{t - l}},{I_{t-s}},{I_{t}}\}$, among which only ${I_{t}}$ is labeled.  	$l$ and $s$ are set  empirically.
		The output is  the parsing result ${P_{t}}$.
	SVP contains three sub-networks.  	As a pre-processing step, we use Faster R-CNN  \cite{ren2015faster} to extract the human region.  Then, the triplet   are fed into Conv1$\sim$Conv5 for discriminative feature extraction.
	The   frame parsing sub-network (Section \ref{sec:image_parser})  produces the rough labelmaps for the triplet, denoted as   $\{ {{\tilde P}_{t - l}},{{\tilde P}_{t-s}},{{\tilde P}_{t}}\}$.
	The optical flow estimation  sub-network aims to estimate the dense correspondence between adjacent frames (Section \ref{sec:optical_flow}).
	The temporal fusion sub-network  (Section \ref{sec:temporal_fusion}) applies the obtained optical flow ${{F_{t,t - l}}}$ and ${{F_{t,t - s}}}$   to ${{\tilde P}_{t - l}}$ and ${{\tilde P}_{t - s}}$, producing ${{\hat P}_{t - l,t}}$ and ${{\hat P}_{t -s,t}}$. To alleviate the influence of  imperfect optical flow, the pixel-wise flow confidences ${{C}_{t - l,t}}$ and ${{C}_{t -s,t}}$ are estimated. The quintet  including
	$\left\{ {{{\tilde P}_t},{{\hat P}_{t - l,t}},{{\hat P}_{t - s,t}},{{\rm{C}}_{{\rm{t-l,t}}}},{{\rm{C}}_{{\rm{t  -  s,t}}}}} \right\}$
	are fused to produce the final ${{P}_t}$, upon which the softmax loss are defined.  Extra supervision is also applied on 	${{\hat P}_{t - l,t}}$ and  ${{\hat P}_{t -s,t}}$ for better performance.
	Our detailed SVP configuration is listed in the supplementary material.

	The image parsing and optical flow estimation  sub-networks share first several  convolution layers because the two tasks are implicitly correlated.
	More specifically,  only pixels with the same labels can be matched by optical flow.
	Besides, both sub-networks
	 make per pixel predictions.
	Frame parsing classifies each pixel while optical flow is the offset/shift of each pixel.
	Therefore, the optimal receptive fields of the  two tasks are similar, which provides a prerequisite  for feature sharing. The other benefit is to save a lot of computation.
	
	\subsection{Frame Parsing Sub-network}  \label{sec:image_parser}
	
	As shown in Figure \ref{fig:framework}, the frame parsing sub-network has three duplications with shared weights to deal with  $I_{t-l}$, $I_{t-s}$ and $I_t$ respectively.
	The input is the  $3$-channel RGB image, and the output is the $K$ channel confidence maps of the same  resolution.
	In our experiments,   DeepLab  \cite{chen2014semantic} is used.
	Our SVP framework is quite generic and is not limited to any specific image parsing method, other semantic segmentation methods \cite{dai2015convolutional,liu2015semantic,liu2015semantic,badrinarayanan2015segnet,long2014fully} can also be used.

	\subsection{Optical Flow Estimation Sub-network}  \label{sec:optical_flow}
	
	We resort to optical flow  ${F_{a,b}}:{R^2} \to {R^2}$  to  build the pixel-wise correspondences between frames. The flow field  ${F_{a,b}^p} = ({q_x} - {p_x},{q_y} - {p_y})$ computes  the relative offsets from each point $p$ in image ${{I_a}}$ to a corresponding point $q$ in image ${{I_b}}$.
	The optical flow estimation sub-network estimates  the flow ${{F_{t,t-l}}} = o\left( {{I_t},{I_{t - l}}} \right)$, where $o\left( {a,b} \right)$ is the operation  of predicting the optical flow from  $a$ to $b$. ${{F_{t,t-s}}}$ is  estimated similarly.
	One feasible approach is to off-line calculate the optical flow via the state-of-the-art methods  \cite{weinzaepfel2013deepflow,brox2011large,brox2011large} and load them into the network during optimization.  It makes training and testing be a multi-stage pipeline, and thus very expensive in space and time. However, SVP computes the optical flow on the fly.

	\textbf{Network architecture:}  After the shared Conv1$\sim$Conv5 layers, a ``correlation layer'' \cite{fischer2015flownet,luo2016efficient,mayer2015large} (denoted as ``Corr'' in Figure \ref{fig:framework})   performs multiplicative patch comparisons between two feature maps.  After that,  several  ``upconvolutional'' layers are introduced to obtain the optical flow with the same resolution as the input image pairs.
	Since our surveillance dataset has no groundtruth optical flow, we use flying chairs dataset \cite{mayer2015large} for training.

	\subsection{Temporal Fusion Sub-network}  \label{sec:temporal_fusion}
	
	\textbf{Optical flow confidence estimation: } The optical flow estimated via the above mentioned method is imperfect. To suppress noisy  ${{\hat P}_{t - l}}$, we estimate the confidence of the estimated optical flow ${{F_{t, t-l}}}$  of each pixel.  The flow ${{F_{t, t-s}}}$ can be handled in similar manners.

	The flow confidence is defined based on the appearance reconstruction criterion \cite{brox2011large}.
	Mathematically, for each pixel $i$ in the optical flow ${{F_{t, t-l}}}$, its  confidence  $C_{t, t-l}^i$ is:
	
	\begin{equation}
	\begin{array}{l}
	C_{t - l,t}^i = {\left\| {I_t^i - \hat I_t^i} \right\|_1}
	= {\left\| {I_t^i - {w^i}\left( {{I_{t - l}},{F_{t, t-l}}} \right)} \right\|_1},\\
	\end{array}
	\label{equ:op_loss}
	\end{equation}
	${\left\| {\cdot} \right\|_1}$ denoted the $L_1$ norm.
	${{\hat I}_t^i}$ is the wrapped counterpart of $I_t^i$.  $w\left( a,b \right)$ is the operation of applying the estimated  optical flow $b$ to warp image $a$.  The coordinates of pixel  $i$ in  ${{I_{t}}}$ is  $\left( {{x^i},{y^i}} \right)$, while the mapped coordinates in  ${{I_{t - l}}}$ is
	$\left( {{x^{i'}},{y^{i'}}} \right) = \left( {{x^i},{y^i}} \right) + F_{t, t-l}^i$. When  $\left( {{x^{i'}},{y^{i'}}} \right)$  falls into  sub-pixel coordinate, we rewrite the $\hat I_t^i$ of Equation \ref{equ:op_loss} via bilinear interpolation:
	\begin{equation}
	\small
	\begin{array}{l}
	\hat I_t^i = {w^i}\left( {{I_{t - l}},{F_{t, t-l}}} \right)\\
	= \sum\limits_{q \in \{ neighbors \; of \; ({x^{i'}},{y^{i'}})\} } {I_{t - l}^q(1 - \left| {{x^{i'}} - {x^q}} \right|)(1 - \left| {{y^{i'}} - {y^q}} \right|)},
	\label{eq:of_map}
	\end{array}
	\end{equation}
	where $q$ denotes the 4-pixel neighbors (top-left, top-right, bottom-left, bottom-right)  of  $\left( {{x^i},{y^i}} \right)$.
	
	The confidence defined in Equation \ref{equ:op_loss} is the distance between the orignal image
	and its warped counterpart. The similaritiy is calcualted via:
	
	\begin{equation}
	C_{t-l,t}^i=exp(-{{C_{t-l,t}^i}}/{{2 \sigma}^2}),
	\end{equation}
	where $\sigma$ is the mean value of $C_{t-l,t}$. Higher value indicates more confident optical flow estimation.

	\textbf{Temporal fusion:}
	As shown in Figure \ref{fig:framework}, the estimated parsing results  $\tilde P_{t-l}$ and $\tilde P_{t-s}$ are  warped according to the optical flow ${F_{t, t-l}}$ and ${F_{t,t-s}}$ via:
	\begin{equation}
	\begin{array}{l}
	{\hat P_{t-l, t}} = w({\tilde P_{t - l}},{F_{t, t-l}}),  \\
	{\hat P_{t-s, t}} = w({\tilde P_{t - s}},{F_{t, t-s}}).
	\end{array}
	\end{equation}

	They are further weighted by the confidence map of Equation \ref{equ:op_loss} to reduce the influence of inaccurate optical flow by:　 ${\hat P_{t-l, t}} \cdot {C_{t - l,t}}$ and ${\hat P_{t-s, t}} \cdot {C_{t - s,t}}$, where  $\cdot$ denotes dot product.  They are fused with ${\tilde P_{t}}$ via  a temporal fusion layer with several $1\times 1$ filters to produce the final ${P_{t}}$.  To enforce accurate model training, we add extra/deep \cite{lee2015deeply} supervision  to ${\hat P_{t-l, t}}$,  ${\hat P_{t-s, t}}$
	and ${P_{t}}$.

	\begin{table*}[t]
		\caption{Per-Class Comparison of F-1 scores  with   state-of-the-arts  and several architectural variants of our model in Indoor dataset.  ($\%$).}
		\label{tabel:each_category_indoor_state}
		\centering
		\small
		\begin{tabular}{p{2cm}|cp{0.6cm}cp{0.6cm}cp{0.6cm}cp{0.6cm}cp{0.6cm}cp{0.6cm}cp{0.6cm}cp{0.6cm}cp{0.6cm}cp{0.6cm}cp{0.6cm}cp{0.6cm}cp{0.6cm}cp{0.6cm}cp{0.6cm}}
			\hline
			\hline
			Methods    & bk	& face	& hair	&  U-clothes	& L-arm	& R-arm	& pants	& L-leg	& R-leg	 	& Dress & L-shoe	& R-shoe	&bag		\\
			\hline
			PaperDoll \cite{yamaguchi2013paper}    &92.62	&57.16	&58.22	&62.52	&19.96	&14.99	&52.47 &25.43	&20.7 &9.92	&20.66	&24.41	&14.32		\\
			\hline
			FCN-8s  \cite{long2014fully}    		&				94.80 & 71.35 & 74.90 & 79.53 & 33.55 & 32.29 & 81.89 & 36.57 & 33.98 & 43.53 & 33.03 & 31.50 & 43.66 \\
			\hline	
			DeepLab	 \cite{chen2014semantic}  	&93.64 & 63.01 & 69.61 & 81.54 & 40.97 & 40.31 & 81.12 & 34.25 & 33.24 & 64.60 & 28.39 & 26.40 & 56.50 \\
			\hline
			EM-Adapt \cite{papandreou2015weakly}     & 93.46 & 66.54 & 70.54 & 77.72 & 42.95 & 42.20 & 82.19 & 39.42 & 37.19 & 63.22 & 33.18 & 31.68 & 53.00 \\
			\hline
			\hline
			SVP l	  	&	94.68 & 67.28 & 72.74 & 82.12 & 42.96 & 43.35 & 81.91 & 39.26 & 38.31 & 67.17 & 31.47 & 30.38 & 58.99 \\
			\hline
			SVP s  	& 	94.65 & 66.27 & 73.48 & 83.12 & 45.17 & 44.89 & 82.72 & 38.62 & 38.43 & 66.04 & 30.93 & 31.46 & 58.81 \\
			\hline
			SVP l+c	  	&	94.44 & 67.29 & 73.76 & 83.06 & 43.56 & 43.56 & 82.33 & 41.36 & 39.46 & 68.36 & 31.75 & 31.73 & 59.04 \\
			\hline
			SVP s+c	  	&	94.64 & 67.62 & 74.13 & 83.48 & 45.13 & \textbf{45.08} & 83.21 & 39.89 & 40.11 & 68.17 & 31.15 & 32.27 & 58.75 \\
			\hline
			SVP l+s  	&	94.50 & 67.08 & 73.52 & 83.10 & \textbf{45.51} & 44.26 & 82.59 & 41.82 & \textbf{42.31} & 69.43 & 33.71 & 33.36 & 58.58 \\
			\hline
			SVP l+s+c	  	&	\textbf{94.89} & \textbf{70.28} & \textbf{76.75} & \textbf{84.18} & 44.79 & 43.29 & \textbf{83.59} & \textbf{42.69} & 40.30 & \textbf{70.76} & \textbf{34.77} & \textbf{35.81} & \textbf{60.43} \\
			\hline
		\end{tabular}
	\end{table*}

	\begin{table*}[t]
		\caption{Per-Class Comparison of F-1 scores  with   state-of-the-arts  and several architectural variants of our model in Outdoor dataset.  ($\%$).}
		\label{tabel:each_category_outdoor_state}
		\centering
		\small
		\begin{tabular}{p{2cm}|cp{0.6cm}cp{0.6cm}cp{0.6cm}cp{0.6cm}cp{0.6cm}cp{0.6cm}cp{0.6cm}cp{0.6cm}cp{0.6cm}cp{0.6cm}cp{0.6cm}cp{0.6cm}cp{0.6cm}cp{0.6cm}}
			\hline
			\hline
			Methods    & bk	& face	& hair	&  U-clothes	& L-arm	& R-arm	& pants	& L-leg	& R-leg	 	 & L-shoe	& R-shoe	&bag		\\
			\hline
			PaperDoll \cite{yamaguchi2013paper}    &87.73	&51.85	&61.50	&66.33	&2.33	&3.39	&22.93 &2.07	&2.70	&17.62	&20.22	&2.45		\\
			\hline
			FCN-8s  \cite{long2014fully}    &92.00 	&
			62.64&	65.58&	78.64	&28.73&	28.97&	79.69&	38.88&	9.08&		32.04&	30.56&	29.45& \\
			\hline	
			DeepLab	 \cite{chen2014semantic}  	&	 92.19 & 58.65 & 66.72 & 84.31 & 42.23 & 35.36 & 81.12 & 30.64 & 6.13 &       37.89 & 33.25 & 52.25 \\
			\hline
			EM-Adapt \cite{papandreou2015weakly}     & 	92.68 & 60.84 & 67.17 & 84.78 & 41.28 & 33.61 & \textbf{81.80} & \textbf{42.39} & 7.28   & 39.54 & 32.20 & \textbf{54.31} \\
			\hline
			\hline
			SVP l	  	&		91.13 & 62.40 & 67.73 & 84.64 & 45.18 & 31.40 & 80.66 & 30.28 & 5.86 &  \textbf{40.32} & 33.11 & 54.96 \\
			\hline
			SVP s  	& 		92.51 & 64.25 & 67.14 & 84.99 & 45.28 & 32.14 & 79.71 & 32.31 & 18.49 &    37.24 & 31.45 & 51.58 \\
			\hline
			SVP l+c	  	&		92.60 & 63.76 & 68.77 & 84.84 & \textbf{45.83} & 33.75 & {81.67} & 31.37 & \textbf{19.06} &       38.54 & 33.51 & 53.57 \\
			\hline
			SVP s+c	  	&		\textbf{92.94} & 64.40 & 69.93 & \textbf{85.43} & 44.44 & 31.86 & 81.65 & 35.88 & 18.22 &        37.48 & 33.36 & {54.23} \\
			\hline
			SVP l+s  	&		91.90 & 63.32 & 69.48 & 84.84 & 42.09 & 28.64 & 80.45 & 31.10 & 13.28 &        38.52 & \textbf{35.52} & 46.89 \\
			\hline
			SVP l+s+c	  	&		92.27 & \textbf{64.49} & \textbf{70.08} & 85.38 & 39.94 & \textbf{35.82} & 80.83 & 30.39 & 13.14   & 37.95 & 34.54 & 50.38 \\
			\hline
		\end{tabular}
	\vspace{-2mm}
	\end{table*}

	\subsection{Training Strategies}
	
	Like the Faster R-CNN \cite{ren2015faster}, we  adopt a  4-step alternating training algorithm for optimization.
	\textbf{(\romannumeral1)} we train the optical flow sub-network  via the strategies in Section  \ref{sec:optical_flow} with flying chairs dataset \cite{fischer2015flownet,mayer2015large}.
	\textbf{(\romannumeral2)} we train the frame parsing sub-network and the temporal fusion sub-network together using the optical flow estimated in step {(\romannumeral1)}.
	Both optical flow and frame parsing sub-networks are initialized with VGG model \cite{simonyan2014very}. The temporal fusion sub-network is initialized via standard Gaussian distribution (with zero mean and unit variance). At this point the two networks do not share convolutional layers.
	\textbf{(\romannumeral3)} We fix the Conv1$\sim$Conv5 layers of optical flow estimation sub-network by those of  frames parsing sub-network,  and only fine-tune the layers unique to optical flow. Now the two sub-networks share convolutional layers.   \textbf{(\romannumeral4)} keeping  Conv1$\sim$Conv5  layers fixed, we fine-tune the unique layers of frame parsing and temporal fusion sub-networks. As such, all sub-networks form a unified network.

	\subsection{Inference}
	During inference, we slide a parsing window along the video to specifically consider the temporal context. The parsing results of $I_t$ is jointly determined by the short video segment preceding
	it.  For calculation simplicity, a triplet of frames, including
	long-range frame $I_{t-l}$,  short-range frame  $I_{t-s}$ as well as $I_t$ collaboratively contribute to the final parsing results $P_t$.
	Note that because the first $l$ frames of a video do not have enough preceding frames to form a sliding parsing window, we apply the frame parsing sub-network alone  to $I_t$  and produce its parsing results.

	\section{Experiments}
	
	\subsection{Experimental setting}
	
	\textbf{Dataset \& Evalutaion Metrics: } Since there is no publicly available surveillance video parsing dataset, we manually build two datasets, one for indoor, the other for outdoor.
	The \textbf{indoor} dataset contains $700$ videos, among which $400$ videos  and  $300$ videos are used as training set and test set, respectively.
	The \textbf{outdoor} dataset contains  $198$ training videos, and $109$ testing videos.
	For both datasets, we randomly select and pixel-wisely label $1$ frame from each training video. For each testing video, we randomly label $5$ frames for
	comprehensive testing.
	The indoor dataset contains $13$ categories, namely
	face,	hair,	upper-clothes,	left-arm,	right-arm,	pants,	left-leg,	right-leg,	left-shoe,	right-shoe,	bag, dress, and background.
	The videos in the outdoor dataset are collected in winter, so the label ``dress'' is missing.
	To obtain human centric video,  human are first detected via Faster R-CNN \cite{ren2015faster}  fine-tuned  on VOC dataset \cite{Everingham15}. The positive samples  are obtained by merging  all foreground pixels.  We enlarge the bounding box horizontally and vertically by $1.2$ times to avoid any missing foreground regions. The obtained human centric images are fed into SVP in both training and testing phases.
	
	We use the same  metric as PaperDoll \cite{yamaguchi2013paper} to  evaluate the performance, including several  standard metrics: accuracy, foreground accuracy,  average precision, average recall and average F-1 score over pixels. Among the five evaluation metrics, the average F-1  is the most important metric.      We train SVP via the Caffe \cite{jia2014caffe} code using Titan X.  The initial learning rates for frame parsing and optical flow estimation sub-networks are 1e-8 and 1e-5 respectively.  In our experiment, the long range $l$ and short range $s$ are empirically  set as $3$ and $1$ in the indoor dataset.  Because the outdoor dataset has a lower frame rate and contains  more quick dynamics, $l$ and $s$ are  set to $2$ and $1$ empirically.

	\begin{table}[htbp]
		\centering
		\small
		\caption{Comparison   with   state-of-the-arts  and several architectural variants of our model in Indoor dataset. ($\%$).}
		\begin{tabular}{c|cccccc}
			\hline
			\hspace{-4mm}		Methods  \hspace{-3mm}&\hspace{-3mm} {Accu}  &\hspace{-3mm}  {fg\_accu}  &\hspace{-3mm}  {Avg.pre}  &\hspace{-3mm}   {Avg.rec}  &\hspace{-3mm}  {Avg. F-1} \\
			\hline
			\hline
			\hspace{-4mm}		PaperDoll  \cite{yamaguchi2013paper} \hspace{-3mm} &\hspace{-3mm} 46.71      &\hspace{-3mm} 78.69      &\hspace{-3mm}33.55      &\hspace{-3mm} 	45.68     &\hspace{-3mm} 36.41 \\
			\hline
			\hspace{-4mm}		FCN-8s  \cite{long2014fully}  \hspace{-3mm}&\hspace{-3mm} 88.33      &\hspace{-3mm} 71.56      &\hspace{-3mm} 55.05      &\hspace{-3mm} 52.15      &\hspace{-3mm} 53.12 \\
			\hline
			\hspace{-4mm}		DeepLab  \cite{chen2014semantic}  \hspace{-3mm}&     86.88 &\hspace{-3mm} 77.45 &\hspace{-3mm} 49.88 &\hspace{-3mm} \textbf{64.30} &\hspace{-3mm}  54.89 \\
			\hline
			\hspace{-4mm}		EM-Adapt  \cite{papandreou2015weakly} \hspace{-3mm} &    86.63 &\hspace{-3mm} 80.88 &\hspace{-3mm} 53.01 &\hspace{-3mm} 63.64 &\hspace{-3mm}  56.40 \\
			\hline
			\hline
			\hspace{-4mm}    	SVP l \hspace{-3mm} &    88.81 &\hspace{-3mm} 74.42 &\hspace{-3mm} 56.28 &\hspace{-3mm} 59.81 &\hspace{-3mm}  57.74 \\
			\hline
			\hspace{-4mm}    	SVP s \hspace{-3mm} &    88.91 &\hspace{-3mm} 77.12 &\hspace{-3mm} 55.90 &\hspace{-3mm} 61.21 &\hspace{-3mm}  58.04 \\
			\hline
			\hspace{-4mm}    	SVP l+c \hspace{-3mm} &    88.75 &\hspace{-3mm} 77.28 &\hspace{-3mm} 56.07 &\hspace{-3mm} 61.94 &\hspace{-3mm}  58.43 \\
			\hline
			\hspace{-4mm}    	SVP s+c \hspace{-3mm} &    89.07 &\hspace{-3mm} 77.06 &\hspace{-3mm} 56.86 &\hspace{-3mm} 61.98 &\hspace{-3mm}  58.73 \\
			\hline
			\hspace{-4mm}    	SVP l+s \hspace{-3mm} &    88.85 &\hspace{-3mm} \textbf{78.68} &\hspace{-3mm} 56.77 &\hspace{-3mm}  {62.73} &\hspace{-3mm}  59.21 \\
			\hline
			\hspace{-4mm}    	SVP l+s+c \hspace{-3mm} &    \textbf{89.88} &\hspace{-3mm} 76.48 &\hspace{-3mm} \textbf{61.52} &\hspace{-3mm} 59.38 &\hspace{-3mm}  \textbf{60.20} \\
			\hline
		\end{tabular}%
		\label{tab:compare_baseline_indoor}
	\end{table}%
	
	\begin{table}[htbp]
		\centering
		\small
		\caption{Comparison   with   state-of-the-arts  and several architectural variants of our model in Outdoor dataset. ($\%$).}
		\begin{tabular}{c|cccccc}
			\hline
			\hspace{-4mm}		Methods  \hspace{-3mm}&\hspace{-3mm} {Accu}  &\hspace{-3mm}  {fg\_accu}  &\hspace{-3mm}  {Avg.pre}  &\hspace{-3mm}   {Avg.rec}  &\hspace{-3mm}  {Avg. F-1} \\
			\hline
			\hline
			\hspace{-4mm}		PaperDoll  \cite{yamaguchi2013paper} \hspace{-3mm} &\hspace{-3mm} 41.48      &\hspace{-3mm} 71.14      &\hspace{-3mm} 30.48      &\hspace{-3mm} 33.11      &\hspace{-3mm} 26.23\\
			\hline			
			\hspace{-4mm}		FCN-8s  \cite{long2014fully}  \hspace{-3mm}&\hspace{-3mm} 82.46      &\hspace{-3mm} 70.70     &\hspace{-3mm} 43.22      &\hspace{-3mm} 50.09      &\hspace{-3mm} 44.33 \\
			\hline
			\hspace{-4mm}		DeepLab  \cite{chen2014semantic}  \hspace{-3mm}&     85.07 &\hspace{-3mm} 78.44 &\hspace{-3mm} 49.87 &\hspace{-3mm} 51.10 &\hspace{-3mm}  47.75 \\
			\hline
			\hspace{-4mm}		EM-Adapt  \cite{papandreou2015weakly} \hspace{-3mm} &    85.82 &\hspace{-3mm} 76.87 &\hspace{-3mm} 50.82 &\hspace{-3mm} 52.98 &\hspace{-3mm} 49.07 \\
			\hline
			\hline
			\hspace{-4mm}    	SVP l \hspace{-3mm} &    84.27 &\hspace{-3mm} \textbf{81.51} &\hspace{-3mm} 47.46 &\hspace{-3mm} 55.31 &\hspace{-3mm}   48.28 \\
			\hline
			\hspace{-4mm}    	SVP s \hspace{-3mm} &    85.83 &\hspace{-3mm} 73.48 &\hspace{-3mm} 53.46 &\hspace{-3mm} 50.63 &\hspace{-3mm}  49.01 \\
			\hline
			\hspace{-4mm}    	SVP l+c \hspace{-3mm} &    85.87 &\hspace{-3mm} 77.37 &\hspace{-3mm} 52.66 &\hspace{-3mm} 52.68 &\hspace{-3mm}   49.79 \\
			\hline
			\hspace{-4mm}    	SVP s+c \hspace{-3mm} &    \textbf{86.30} &\hspace{-3mm} 77.13 &\hspace{-3mm} 52.89 &\hspace{-3mm} \textbf{52.70} &\hspace{-3mm}   49.99 \\
			\hline
			\hspace{-4mm}    	SVP l+s \hspace{-3mm} &     85.30 &\hspace{-3mm} 77.03 &\hspace{-3mm} 56.15 &\hspace{-3mm} 49.92 &\hspace{-3mm}   51.17 \\
			\hline
			\hspace{-4mm}    	SVP l+s+c \hspace{-3mm} &    85.71 &\hspace{-3mm} 79.26 &\hspace{-3mm} \textbf{56.95} &\hspace{-3mm} 52.14 &\hspace{-3mm}   \textbf{52.94} \\
			
			\hline
		\end{tabular}%
		\label{tab:compare_baseline_outdoor}
		\vspace{-3mm}
	\end{table}%

	\subsection{Comparison with state-of-the-art}

	We compare our results with five state-of-the-art methods.  The first is  \textbf{PaperDoll} \cite{yamaguchi2013paper}. It is the best traditional method which does not use deep learning.
	The second is \textbf{FCN-8s} \cite{long2014fully}, which achieves competitive results  in several semantic segmentation benchmark datasets.
	We implement the FCN-8s because of its superior performance than other FCN variants, such as FCN-16s and FCN-32s.
	The third baseline is \textbf{DeepLab} \cite{chen2014semantic}.
	The above mentioned  three methods are supervised algorithms.  Therefore, we only use the labeled set for training. The fourth baseline method is
	\textbf{EM-Adapt}  \cite{papandreou2015weakly} which can use both  image-level and  bounding-box annotation as weak- and semi-Supervised supervision.
	We also try another baseline  {DecoupledNet} \cite{hong2015decoupled},   which
	targets for semi-supervised semantic segmentation with  image-level class labels and/or pixel-wise segmentation annotations. However, the results of DecoupledNet in both datasets are much lower than SVP and other baselines. The reason is that DecoupledNet  first obtains the saliency map of each classified label.  Deconvlution is then operated upon the map to generate the final parsing results. However, many labels, e.g., face, hair etc, appear in almost every training image,  which causes the classification network less sensitive to the position  of these labels.	For EM-Adapt and DecoupledNet, we use their source codes   \footnote{http://liangchiehchen.com/projects/DeepLab-LargeFOV-Semi-EM-Fixed.html}
	\footnote{http://cvlab.postech.ac.kr/research/decouplednet/}.

	Table \ref{tab:compare_baseline_indoor} shows  the comparisons between SVP and $4$ state-of-the-art methods in the  \textbf{Indoor}  dataset.
	Different variants of SVP are generated by gradually adding more components, which will be  discussed in the next subsection. It can be seen that our best SVP, namely ``SVP l+s+c'' reaches the average F-1  score of $0.6020$ while PaperDoll, FCN-8s and DeepLab only reach $0.3641$, $0.5312$ and $0.5489$ respectively. In other words, our results are superior than PaperDoll, FCN-8s and DeepLab  by  $0.2379$, $0.0708$ and $0.0531$.  The three baselines only use labeled images. Therefore, the improvements show the advantage of utilizing the unlabeled dataset. EM-Adapt also uses unlabeled images, and thus reaches a higher F1-score of $0.5640$, which is better than the three supervised baselines. However, EM-Adapt is still worse than all the variants of SVP. It shows that label propagation via optical flow is helpful in the surveillance video parsing task. The F1-scores of each category are shown in Table \ref{tabel:each_category_indoor_state}. We can observe that ``SVP l+s+c'' beats  PaperDoll, FCN,  DeepLab and EM-Adapt in all $13$ categories, which again shows the big improvements brought by the proposed SVP framework.

	Table \ref{tab:compare_baseline_outdoor} shows the results of SVP and $4$ state-of-the-art in the  \textbf{Outdoor}  dataset.   It can be seen that our method reaches the average F-1  score of $0.5294$ while PaperDoll, FCS-8s,  DeepLab and EM-Adapt only reach $0.2623$, $0.4433$ and $0.4775$. The improves are
	$0.2671$, $0.0861$ and $0.0519$ respectively.
	Comparing Table \ref{tab:compare_baseline_outdoor} and Table \ref{tab:compare_baseline_indoor}, we find that the performances of all algorithms generally drop. The reason is that the outdoor dataset contains  $198$ training videos,  while the number is doubled in the indoor dataset, reaching $400$. The F1-scores of each category  are shown in Table  \ref{tabel:each_category_outdoor_state}. 	We can observe that ``SVP l+s+c'' beats  PaperDoll, FCN and DeepLab in all $13$ categories and is better than EM-Adapt in most categories, which again shows the effectiveness.

	\subsection{Component Analysis}

\begin{figure}[h]
		\begin{center}
			\includegraphics[width=0.9\linewidth]{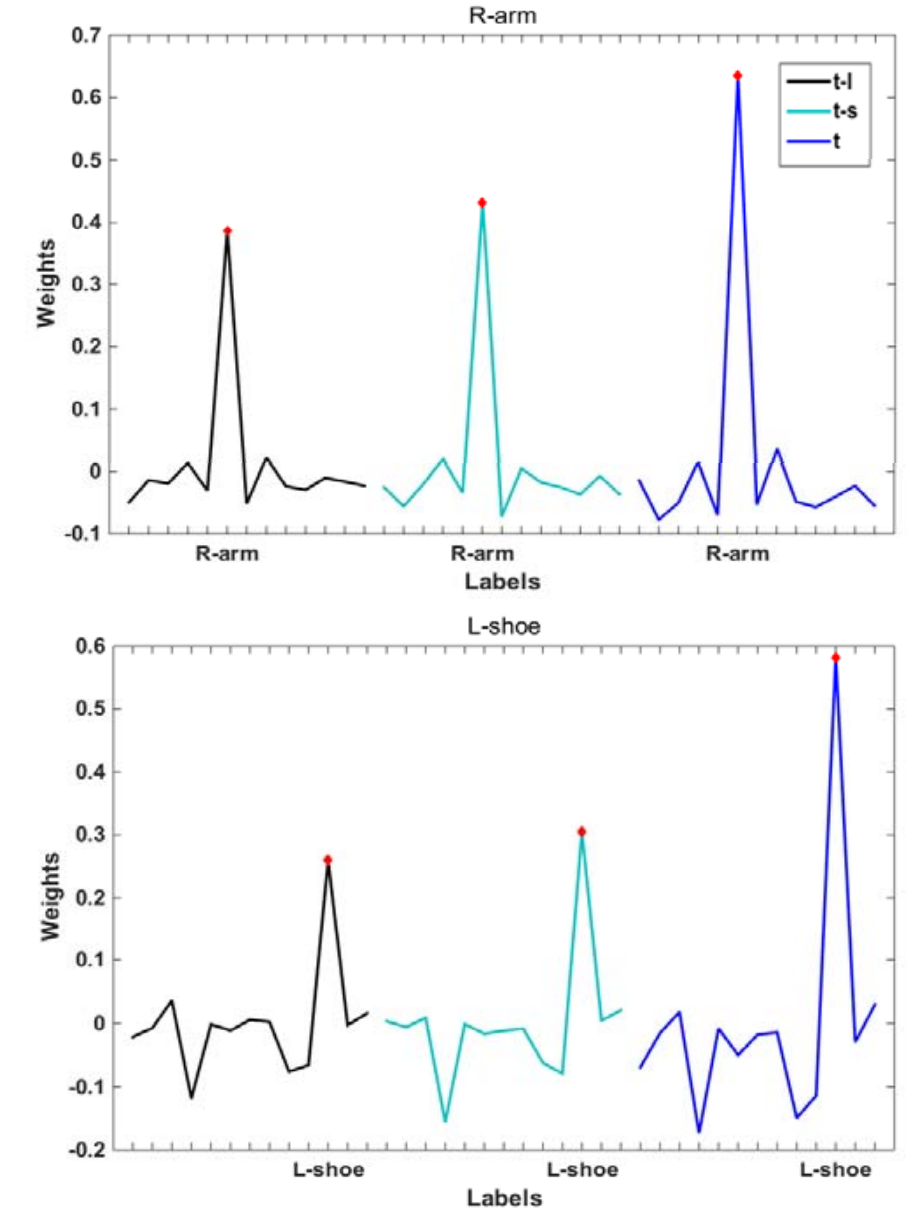}
		\end{center}
			\vspace{-4mm}
		\caption{The temporal pooling weight for R-arm and L-shoe.  }
		\label{fig:temporal_pooling_weight}
					\vspace{-3mm}
	\end{figure}

	\textbf{Temporal fusion weights:} We visualize the learned weights for the temporal fusion layers for R-arm  and L-shoe in Figure \ref{fig:temporal_pooling_weight} in the Indoor dataset. 	The horizontal axis has  $3\times K$ ticks, corresponding to the  $K$  labels for 	 $I_{t-l}$ (shown in black), $I_{t-s}$ (shown in green) and $I_{t}$ (shown in blue) sequentially. The vertical axis illustrates the fusion weights.
	
	By analyzing the sub-figure for R-arm, we have several observations. First,  the shapes of the weights for $I_{t-l}$, $I_{t-s}$ and $I_{t}$ are similar. Second, all maximum values for the triplet (denoted as red dots) are positive, which demonstrates that all frames contribute to the final result.     Third, for all the frames, the labels reaching maximum values  are all  R-arm. Fourth, the maximum value of $I_{t-s}$  is higher than that of $I_{t-l}$, 	because it contains less errors in optical flow. The  maximum value of $I_{t}$  is the highest, because it is the frame under consideration.  Similar phenomenon can be found in the L-shoe case.

	\begin{figure*}[t]
		\begin{center}
			\includegraphics[width=1\linewidth]{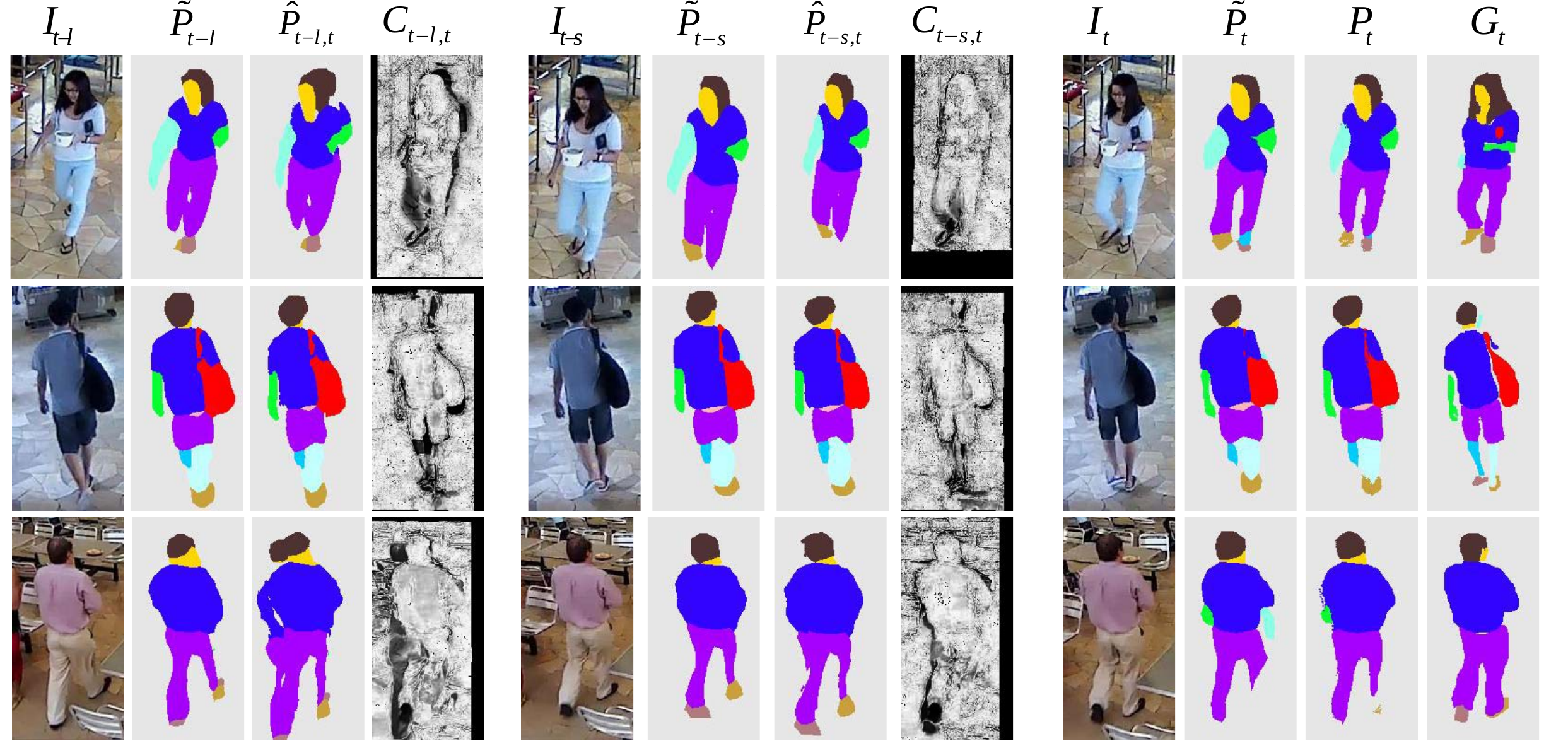}
		\end{center}
	\vspace{-4mm}
		\caption{Step by step illustration of SVP. 1$\sim$4 columns: the long-range frame, its the parsing result, the warped parsing result and the confidence map. 5$\sim$8 columns: the short-range frame, its parsing result, the warped parsing result and the confidence map. 9$\sim$12 columns: test image, the rough parsing result, refined parsing result and ground truth parsing result. }
		\label{fig:of_results}
		\vspace{-4mm}
	\end{figure*}

	\textbf{Long/Short range context: } We test the effectiveness  of long and short range frame.  ``SVP l'' means SVP with long-range context only. To implement this SVP variant, an image pair, namely $I_t$ as well as $I_{t-l}$ are fed into SVP during both training and testing phases.    Similarly, ``SVP s'' is SVP containing only short-range frame. ``SVP l+s'' is the combination of them, meaning both long-range and short-range frames are considered. Table \ref{tab:compare_baseline_indoor} shows the results in indoor dataset. The Ave.F-1 of ``SVP l'' and ``SVP s'' reach $0.5774$ and $0.5804$ respectively, which are lower than ``SVP l+s'' $0.5843$. It proves the   long and short range context are complementary. Similar conclusion can be drawn from outdoor dataset in Table \ref{tab:compare_baseline_outdoor}. ``SVP l'' and ``SVP s'' achieve $0.4828$ and $0.4901$, while the combination of them reaches $0.4979$. The per-class F1 score of ``SVP l'', ``SVP s'' and ``SVP l+s'' in indoor and outdoor datasets  can be found in Table \ref{tabel:each_category_indoor_state} and Table \ref{tabel:each_category_outdoor_state} respectively. They again show that both long and short range context are necessary.

	\begin{figure*}[t]
		\begin{center}
			\includegraphics[width=0.95\linewidth]{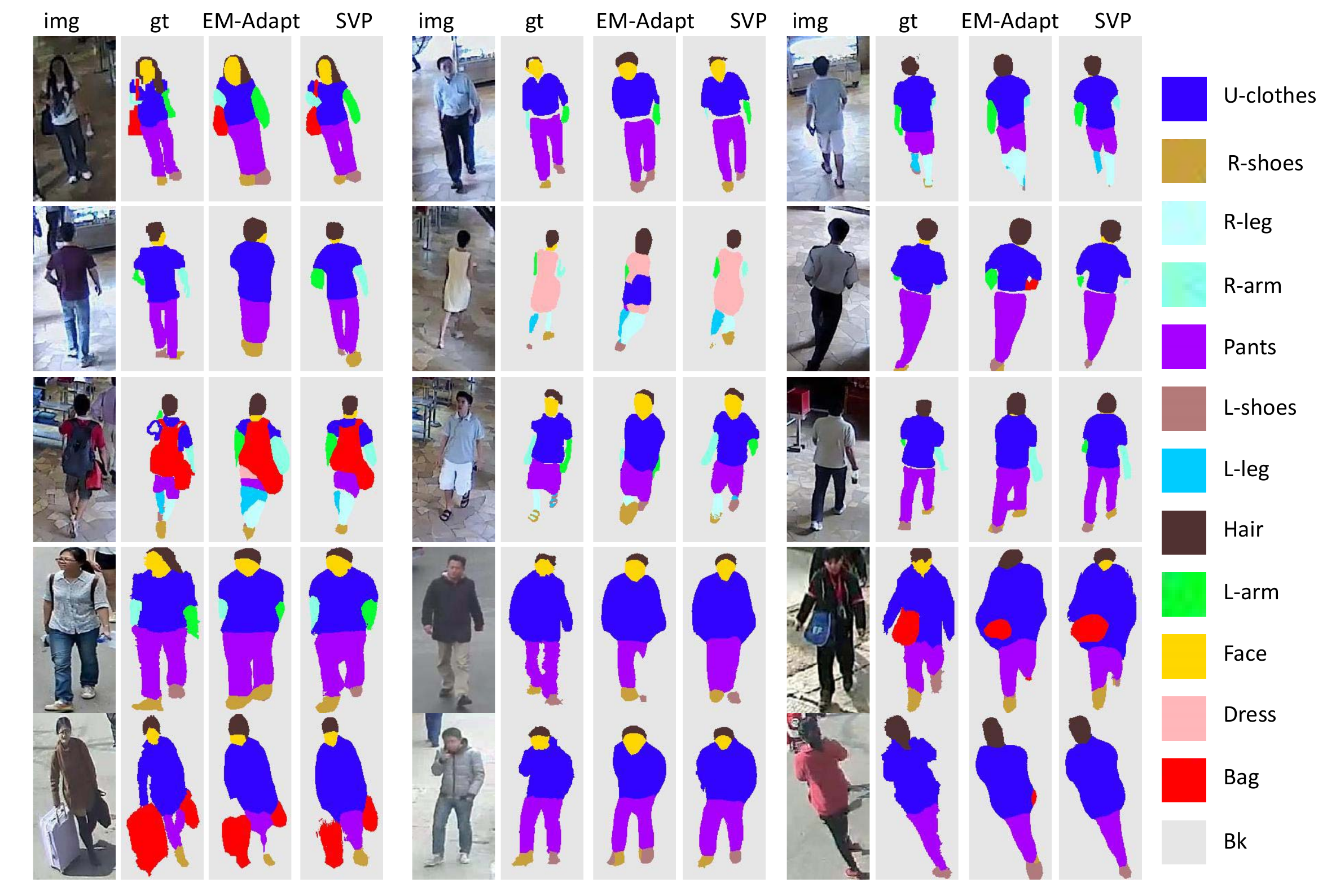}
		\end{center}
			\vspace{-5mm}
		\caption{The test image, the groundtruth  label,  results of the EM-Adapt and SVP are shown sequentially. }
		\label{fig:parsing_show}
		\vspace{-4mm}
	\end{figure*}

	\textbf{Optical flow confidence: } The flow confidence is designed for filtering/suppressing the noisy optical flow.  To this end, we implement two SVP variants called ``SVP l+c'' and ``SVP s+c''  indicating either long or short-range optical flow is weighed by its confidence first and then contribute to the final parsing result. The results in indoor dataset is shown in Table \ref{tab:compare_baseline_indoor}.  We find that ``SVP l+c''  improves ``SVP l'' 	and ``SVP s+c'' performs better than ``SVP s''. This demonstrates the effectiveness of optical  flow confidence. The same conclusion can be drawn by comparing the F-1 score of ``SVP l+s+c'' and ``SVP l+s''.  We also  validate the effectiveness  of optical flow confidence in outdoor dataset. As shown in Table \ref{tab:compare_baseline_outdoor}, the F-1 score of ``SVP l+s+c''  is $0.5294$,  which is  higher than ``SVP l+s'' $0.5117$.
	
	\subsection{Qualitative Results}

	Figure \ref{fig:of_results} shows the stepwise results of  SVP  in indoor dataset. In the first  row,   the left shoe of the women is predicted as leg in ${{\tilde P}_t}$. The warped label from the  $I_{t-s}$, denoted as  ${{\hat P}_{t - s,t}}$ does not find left shoe. Thanks to the fusion from  ${{\hat P}_{t - l,t}}$, the women's left shoe is labelled correctly in the final prediction  ${{P}_{t}}$.
	Again in the first row, comparing with $I_{t-s}$,  the women is far from the camera  in $I_t$, and thus is relatively small. The foreground region
	shrinks  from $\tilde P_{t-s}$   to  ${{\hat P}_{t - s,t}}$,  which shows that the estimated optical flow is very accurate. 	
	Inaccurate optical flow may result in the bad propagated parsing result, e.g., the shape of the hair in  ${{\hat P}_{t - l,t}}$ is too large in the first row. However, the inaccurate hair region has a low confidence in $C_{t-l, t}$. Therefore, the fused result $P_t$ has precise  hair shape. 	
	In the second row, the strap of the bag is almost ignored in ${{\tilde P}_t}$. However, both     ${{\tilde P}_{t-l}}$ and     ${{\tilde P}_{t-s}}$ find the strap, and help to distinguish the strap from the upper-clothes successfully in  ${{P}_t}$. In the third row, the ${{P}_{t}}$ correctly removes the wrongly predicted arm  in ${{\tilde P}_t}$. The  $I_{t-l}$ is not warped very good, and there is a ghost behind this man in the labelmap  ${\hat P_{t - l,t}}$. But fortunately it does not affect the fused prediction $P_t$, because the confidence of this ghost is very low  in ${C_{t - l,t}}$ and hence it is filtered out during the fusion.

	Several qualitative results of both datasets are shown in Figure \ref{fig:parsing_show}.
	The first three rows  show paring results of the indoor dataset while the last two rows demonstrate those of outdoor dataset.
	In each group, the test image, the groundtruth,  the parsing results of EM-Adapt and SVP are shown. It can be seen that SVP is generally better than EM-Adapt from two aspects.  First, SVP correctly estimates the existence of a label. 	For example, for the image in the second row second column, the region wrongly predicted as upper clothes by EM-Adapt is correctly predicted as dress by SVP.
	Another example is second row first column.   EM-Adapt misses the left shoe.  SVP correctly predicts the left shoe's existence and location.
	Second, SVP can better estimate the shape of the labels. For example, in the first image in top row,  the shape of  the bag strap is slender, which is correctly estimated by SVP. Moreover, the  shapes of shoes estimated by SVP are more accurate than EM-Adapt.
    For another example, SVP better identifies the shapes of pants and left/right arms in the third image of the third row.

	\subsection{Time Complexity}
	
	Note that in the inference stage,  much computation can be saved.  For example,  when parsing frame $I_t$,  the long-range frame $I_{t-l}$ and  short-range frame $I_{t-s}$  do not need go through the frame parsing sub-network because their rough parsing results $P_{t-l}$ and $P_{t-s}$ have already been calculated. For another example, the extra computation brought by the optical flow estimation sub-network is small because the Conv1$\sim$Conv5 features are shared.  Moreover, the fusion layer contains  several  $1\times1$   convolutions and thus is not quite time-consuming.
	
	We compare SVP with another comparison method. It has similar network architecture as SVP, except that the optical flow is calculated offline via  EpicFlow \cite{revaud2015epicflow}. The F1 score is $0.6024$ and $0.5311$ in indoor and outdoor dataset respectively, which are slightly better than SVP. However, computing Epicflow needs several seconds' CPU computation while SVP runs on GPU for milliseconds. Also Epicflow should be calculated beforehand. To the contrary, SVP is end-to-end and convenient for real application.

	\section{Conclusion \& Future Works}
	
	In this work, we present an end-to-end single frame supervised video parsing network. To parse a testing frame, SVP processes a video segment preceding it.  The rough frame parsing results and  the on-line computed optical flows among frames are fused to produce  refined parsing results. We demonstrate the effectiveness of SVP on two newly collected surveillance video parsing datasets.
	
	In future, we will build an online demo to parse any surveillance video uploaded by users  in real time. 	Moreover, we plan to apply SVP to parse other kinds of videos, such as urban scene videos \cite{cordts2016cityscapes}.  Besides, we consider  to explore how to further reduce the error of optical flow estimation by considering cycle consistency constraints \cite{zhou2016learning}.

{\small
\bibliographystyle{ieee}
\bibliography{egbib}

\begin{thebibliography}{10}\itemsep=-1pt

\bibitem{badrinarayanan2015segnet}
V.~Badrinarayanan, A.~Handa, and R.~Cipolla.
\newblock Segnet: A deep convolutional encoder-decoder architecture for robust
  semantic pixel-wise labelling.
\newblock {\em arXiv:1505.07293}, 2015.

\bibitem{bai2016exploiting}
M.~Bai, W.~Luo, K.~Kundu, and R.~Urtasun.
\newblock Exploiting semantic information and deep matching for optical flow.
\newblock In {\em ECCV}, 2016.

\bibitem{bearman2015s}
A.~Bearman, O.~Russakovsky, V.~Ferrari, and L.~Fei-Fei.
\newblock What's the point: Semantic segmentation with point supervision.
\newblock {\em arXiv:1506.02106}, 2015.

\bibitem{brox2011large}
T.~Brox and J.~Malik.
\newblock Large displacement optical flow: descriptor matching in variational
  motion estimation.
\newblock {\em TPAMI}, 2011.

\bibitem{chen2014semantic}
L.-C. Chen, G.~Papandreou, I.~Kokkinos, K.~Murphy, and A.~L. Yuille.
\newblock Semantic image segmentation with deep convolutional nets and fully
  connected crfs.
\newblock {\em arXiv:1412.7062}, 2014.

\bibitem{cordts2016cityscapes}
M.~Cordts, M.~Omran, S.~Ramos, T.~Rehfeld, M.~Enzweiler, R.~Benenson,
  U.~Franke, S.~Roth, and B.~Schiele.
\newblock The cityscapes dataset for semantic urban scene understanding.
\newblock {\em arXiv:1604.01685}, 2016.

\bibitem{dai2015boxsup}
J.~Dai, K.~He, and J.~Sun.
\newblock Boxsup: Exploiting bounding boxes to supervise convolutional networks
  for semantic segmentation.
\newblock In {\em ICCV}, 2015.

\bibitem{dai2015convolutional}
J.~Dai, K.~He, and J.~Sun.
\newblock Convolutional feature masking for joint object and stuff
  segmentation.
\newblock In {\em CVPR}, 2015.

\bibitem{deng2014pedestrian}
Y.~Deng, P.~Luo, C.~C. Loy, and X.~Tang.
\newblock Pedestrian attribute recognition at far distance.
\newblock In {\em ACM MM}, 2014.

\bibitem{Everingham15}
M.~Everingham, S.~M.~A. Eslami, L.~Van~Gool, C.~K.~I. Williams, J.~Winn, and
  A.~Zisserman.
\newblock The pascal visual object classes challenge: A retrospective.
\newblock {\em IJCV}, 2015.

\bibitem{fischer2015flownet}
P.~Fischer, A.~Dosovitskiy, E.~Ilg, P.~H{\"a}usser, C.~Haz{\i}rba{\c{s}},
  V.~Golkov, P.~van~der Smagt, D.~Cremers, and T.~Brox.
\newblock Flownet: Learning optical flow with convolutional networks.
\newblock {\em arXiv:1504.06852}, 2015.

\bibitem{fragkiadaki2015learning}
K.~Fragkiadaki, P.~Arbel{\'a}ez, P.~Felsen, and J.~Malik.
\newblock Learning to segment moving objects in videos.
\newblock In {\em CVPR}, 2015.

\bibitem{hong2015decoupled}
S.~Hong, H.~Noh, and B.~Han.
\newblock Decoupled deep neural network for semi-supervised semantic
  segmentation.
\newblock In {\em NIPS}, 2015.

\bibitem{jia2014caffe}
Y.~Jia, E.~Shelhamer, J.~Donahue, S.~Karayev, J.~Long, R.~Girshick,
  S.~Guadarrama, and T.~Darrell.
\newblock Caffe: Convolutional architecture for fast feature embedding.
\newblock {\em arXiv:1408.5093}, 2014.

\bibitem{lee2015deeply}
C.-Y. Lee, S.~Xie, P.~Gallagher, Z.~Zhang, and Z.~Tu.
\newblock Deeply-supervised nets.
\newblock In {\em AISTATS}, volume~2, page~6, 2015.

\bibitem{li2014deepreid}
W.~Li, R.~Zhao, T.~Xiao, and X.~Wang.
\newblock Deepreid: Deep filter pairing neural network for person
  re-identification.
\newblock In {\em CVPR}, 2014.

\bibitem{liang2015deep}
X.~Liang, S.~Liu, X.~Shen, J.~Yang, L.~Liu, L.~Lin, and S.~Yan.
\newblock Deep human parsing with active template regression.
\newblock {\em TPAMI}, 2015.

\bibitem{xiaodaniccv}
X.~Liang, C.~Xu, X.~Shen, J.~Yang, S.~Liu, J.~Tang, L.~Lin, and S.~Yan.
\newblock Human parsing with contextualized convolutional neural network.
\newblock {\em ICCV}, 2015.

\bibitem{lin2016scribblesup}
D.~Lin, J.~Dai, J.~Jia, K.~He, and J.~Sun.
\newblock Scribblesup: Scribble-supervised convolutional networks for semantic
  segmentation.
\newblock {\em arXiv:1604.05144}, 2016.

\bibitem{liu2015semantic}
Z.~Liu, X.~Li, P.~Luo, C.-C. Loy, and X.~Tang.
\newblock Semantic image segmentation via deep parsing network.
\newblock In {\em ICCV}, 2015.

\bibitem{long2014fully}
J.~Long, E.~Shelhamer, and T.~Darrell.
\newblock Fully convolutional networks for semantic segmentation.
\newblock {\em arXiv:1411.4038}, 2014.

\bibitem{luo2013pedestrian}
P.~Luo, X.~Wang, and X.~Tang.
\newblock Pedestrian parsing via deep decompositional network.
\newblock In {\em ICCV}, 2013.

\bibitem{luo2016efficient}
W.~Luo, A.~G. Schwing, and R.~Urtasun.
\newblock Efficient deep learning for stereo matching.
\newblock In {\em CVPR}, 2016.

\bibitem{mayer2015large}
N.~Mayer, E.~Ilg, P.~H{\"a}usser, P.~Fischer, D.~Cremers, A.~Dosovitskiy, and
  T.~Brox.
\newblock A large dataset to train convolutional networks for disparity,
  optical flow, and scene flow estimation.
\newblock {\em arXiv:1512.02134}, 2015.

\bibitem{noh2015learning}
H.~Noh, S.~Hong, and B.~Han.
\newblock Learning deconvolution network for semantic segmentation.
\newblock In {\em ICCV}, 2015.

\bibitem{papandreou2015weakly}
G.~Papandreou, L.-C. Chen, K.~P. Murphy, and A.~L. Yuille.
\newblock Weakly-and semi-supervised learning of a deep convolutional network
  for semantic image segmentation.
\newblock In {\em ICCV}, 2015.

\bibitem{pfister2015flowing}
T.~Pfister, J.~Charles, and A.~Zisserman.
\newblock Flowing convnets for human pose estimation in videos.
\newblock {\em arXiv:1506.02897}, 2015.

\bibitem{ren2015faster}
S.~Ren, K.~He, R.~Girshick, and J.~Sun.
\newblock Faster r-cnn: Towards real-time object detection with region proposal
  networks.
\newblock In {\em NIPS}, pages 91--99, 2015.

\bibitem{revaud2015epicflow}
J.~Revaud, P.~Weinzaepfel, Z.~Harchaoui, and C.~Schmid.
\newblock Epicflow: Edge-preserving interpolation of correspondences for
  optical flow.
\newblock In {\em CVPR}, 2015.

\bibitem{sevilla2016optical}
L.~Sevilla-Lara, D.~Sun, V.~Jampani, and M.~J. Black.
\newblock Optical flow with semantic segmentation and localized layers.
\newblock {\em arXiv:1603.03911}, 2016.

\bibitem{shelhamer2016clockwork}
E.~Shelhamer, K.~Rakelly, J.~Hoffman, and T.~Darrell.
\newblock Clockwork convnets for video semantic segmentation.
\newblock {\em arXiv:1608.03609}, 2016.

\bibitem{simonyan2014very}
K.~Simonyan and A.~Zisserman.
\newblock Very deep convolutional networks for large-scale image recognition.
\newblock {\em arXiv:1409.1556}, 2014.

\bibitem{wang2015joint}
P.~Wang, X.~Shen, Z.~Lin, S.~Cohen, B.~Price, and A.~L. Yuille.
\newblock Joint object and part segmentation using deep learned potentials.
\newblock In {\em ICCV}, pages 1573--1581, 2015.

\bibitem{weinzaepfel2013deepflow}
P.~Weinzaepfel, J.~Revaud, Z.~Harchaoui, and C.~Schmid.
\newblock Deepflow: Large displacement optical flow with deep matching.
\newblock In {\em ICCV}, 2013.

\bibitem{xia2015zoom}
F.~Xia, P.~Wang, L.-C. Chen, and A.~L. Yuille.
\newblock Zoom better to see clearer: Human part segmentation with auto zoom
  net.
\newblock {\em arXiv:1511.06881}, 2015.

\bibitem{xu2015learning}
J.~Xu, A.~G. Schwing, and R.~Urtasun.
\newblock Learning to segment under various forms of weak supervision.
\newblock In {\em CVPR}, 2015.

\bibitem{yamaguchi2013paper}
K.~Yamaguchi, M.~H. Kiapour, and T.~Berg.
\newblock Paper doll parsing: Retrieving similar styles to parse clothing
  items.
\newblock In {\em ICCV}, 2013.

\bibitem{zhao2014learning}
R.~Zhao, W.~Ouyang, and X.~Wang.
\newblock Learning mid-level filters for person re-identification.
\newblock In {\em CVPR}, 2014.

\bibitem{zhou2016learning}
T.~Zhou, P.~Kr{\"a}henb{\"u}hl, M.~Aubry, Q.~Huang, and A.~A. Efros.
\newblock Learning dense correspondence via 3d-guided cycle consistency.
\newblock {\em arXiv:1604.05383}, 2016.

\end{thebibliography}
}

\end{document}